\documentclass[sigconf]{acmart}
\usepackage[show]{chato-notes}

\newcommand{\yu}[1]{\textcolor{black}{#1}}

\newcommand{\red}[1]{\textcolor{black}{#1}}

\newcommand{\yuold}[1]{\textcolor{black}{#1}}
\newcommand{\wuold}[1]{\textcolor{black}{#1}}

\usepackage{graphicx}
\usepackage{amsfonts}
\usepackage{subfigure}
\usepackage{subcaption}
\usepackage{multirow}
\usepackage{multicol}   
\usepackage{dashrule}   
\usepackage[para]{footmisc}
\usepackage{graphics}
\usepackage{soul}
\usepackage{float}
\usepackage{tcolorbox}
\usepackage{enumitem}

\AtBeginDocument{%
  }

\setcopyright{acmlicensed}
\copyrightyear{2025}
\acmYear{2025}
\acmDOI{XXXXXXX.XXXXXXX}

\begin{document}

\title{Thought-Augmented Planning for LLM-Powered Interactive Recommender Agent}

\author{Haocheng Yu}
\affiliation{%
    \institution{University of Science and Technology of China}
    \city{Hefei}
    \state{Anhui}
    \country{China}
}
\email{yuhaoch@mail.ustc.edu.cn}

\author{Yaxiong Wu}
\affiliation{%
    \institution{Huawei Noah’s Ark Lab}
    \city{Shenzhen}
    \state{Guangdong}
    \country{China}
}
\email{wu.yaxiong@huawei.com}

\author{Hao Wang}
\affiliation{%
    \institution{University of Science and Technology of China}
    \city{Hefei}
    \state{Anhui}
    \country{China}
}
\email{wanghao3@ustc.edu.cn}

\author{Wei Guo}
\affiliation{%
    \institution{Huawei Noah’s Ark Lab}
    \city{Shenzhen}
    \state{Guangdong}
    \country{China}
}
\email{guowei67@huawei.com}

\author{Yong Liu}
\affiliation{%
    \institution{Huawei Noah’s Ark Lab}
    \city{Shenzhen}
    \state{Guangdong}
    \country{China}
}
\email{liu.yong6@huawei.com}

\author{Yawen Li}
\affiliation{%
    \institution{Beijing University of Posts and Telecommunications}
    \city{Beijing}
    \country{China}
}
\email{warmly0716@126.com}

\author{Yuyang Ye}
\affiliation{%
    \institution{University of Science and Technology of China}
    \city{Hefei}
    \state{Anhui}
    \country{China}
}
\email{yeyuyang@mail.ustc.edu.cn}

\author{Junping Du}
\affiliation{%
    \institution{Beijing University of Posts and Telecommunications}
    \city{Beijing}
    \country{China}
}
\email{junpingd@bupt.edu.cn}

\author{Enhong Chen}
\affiliation{%
    \institution{University of Science and Technology of China}
    \city{Hefei}
    \state{Anhui}
    \country{China}
}
\email{cheneh@ustc.edu.cn}

\begin{abstract}

Interactive recommendation is a typical information-seeking task that allows users to interactively express their needs through natural language and obtain personalized recommendations.
Large language model-powered (LLM-powered) agents have become a new paradigm in interactive recommendations, effectively capturing users' real-time needs and enhancing personalized experiences. 
\yu{However, due to limited planning and generalization capabilities, existing formulations of LLM-powered interactive recommender agents struggle to effectively address diverse and complex user intents, such as intuitive, unrefined, or occasionally ambiguous requests. 
To tackle this challenge, we propose a novel thought-augmented interactive recommender agent system (TAIRA) that addresses complex user intents through distilled thought patterns. 
Specifically, TAIRA is designed as an LLM-powered multi-agent system featuring a manager agent that orchestrates recommendation tasks by decomposing user needs and planning subtasks, with its planning capacity strengthened through Thought Pattern Distillation (TPD), a thought-augmentation method that extracts high-level thoughts from the agent's and human experts' experiences. 
Moreover, we designed a set of user simulation schemes to generate personalized queries of different difficulties and evaluate the recommendations based on specific datasets. Through comprehensive experiments conducted across multiple datasets, TAIRA exhibits significantly enhanced performance compared to existing methods. Notably, TAIRA shows a greater advantage on more challenging tasks while generalizing effectively on novel tasks, further validating its superiority in managing complex user intents within interactive recommendation systems. }
The code is publicly available at: \url{https://github.com/Alcein/TAIRA}.\par

\end{abstract}


\fancyhead[]{}

\maketitle

\begin{figure}[t]
    \centering
    \includegraphics[width=\linewidth]{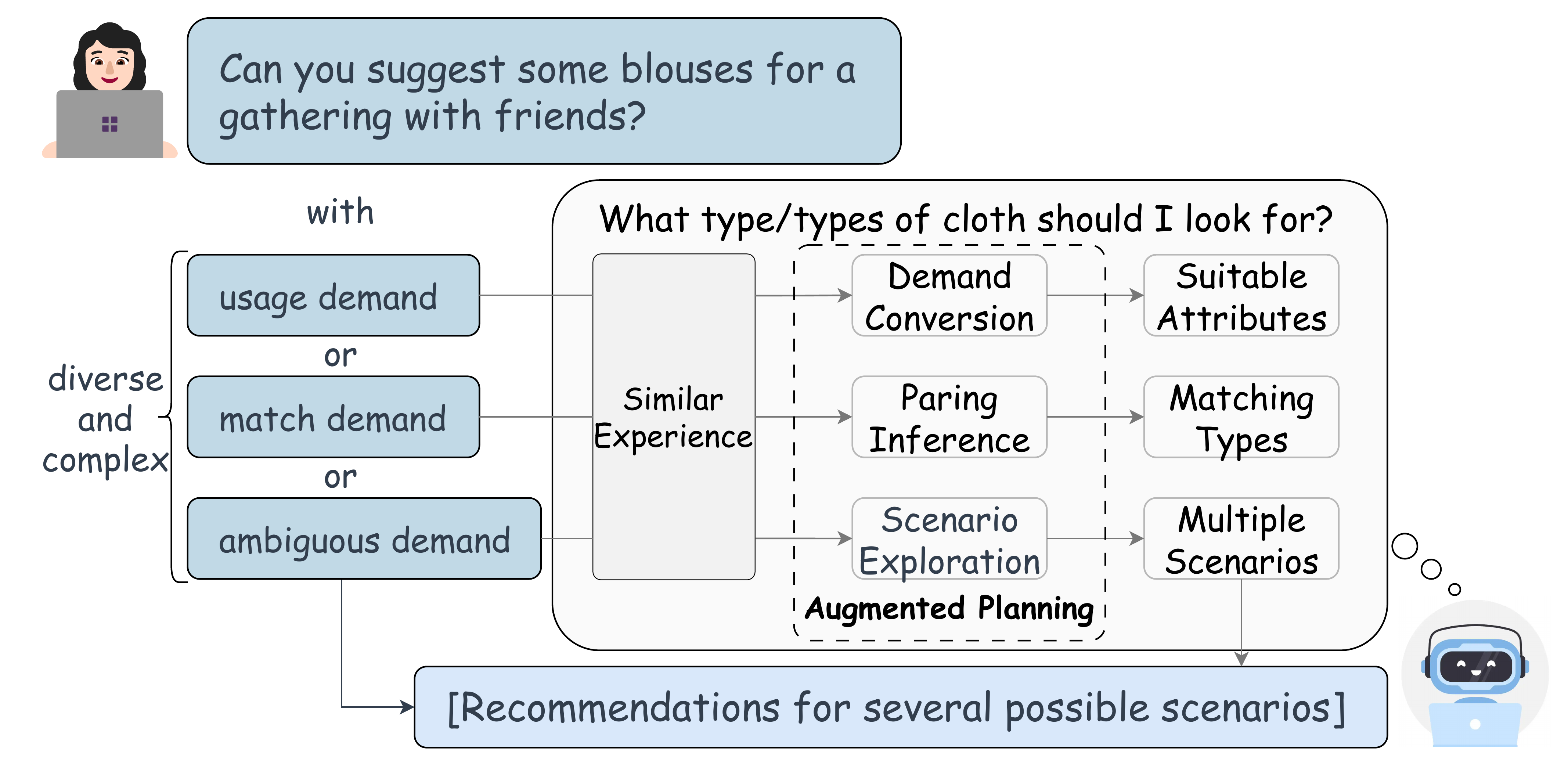}
    \vspace{-1.\baselineskip}
    \caption{Examples of recommendation involving diverse and complex user intent and thought-augmentation with past experiences.}
    \label{fig:introduction} \vspace{-1.\baselineskip}
\end{figure}

\section{Introduction}

Interactive Recommender Systems (IRS) mark a significant advancement in recommendation. Unlike traditional systems, IRS allows users to actively communicate their needs through natural language, allowing the system to accurately capture immediate preferences. This interactive approach not only ensures personalized recommendations but also heightens user engagement and satisfaction \cite{HE20169}. Traditional IRS relies on predefined rules and templates, guiding users through questions about item attributes to refine recommendations~\cite{lei2020estimation}. With advancements in natural language processing, especially the rise of \yuold{LLM-powered agents,} IRS has significantly improved in dialogue understanding and generation \cite{huang2024recommenderaiagentintegrating}, enabling more natural user interactions and greater accuracy in completing recommendation tasks.\par

\yu{Despite these advancements, existing agents for interactive recommendation~\cite{huang2024recommenderaiagentintegrating} still face challenges when handling complex and diverse user intents. To expand the adoption of interactive recommender systems to a broader user base with varied engagement patterns and address potential challenges, the system's tasks will grow more complex, and users' queries will become increasingly diverse. As shown in Figure \ref{fig:introduction}, some requests focus on usage scenarios without specifying item details (e.g., "blouses for a gathering with friends"), while others require product combinations or present ambiguous demands. 
Users expect recommendation systems to act as consultants, capable of understanding complex intentions and translating them into meaningful products. This presents a significant challenge in recommendation scenarios, as it demands advanced reasoning and planning abilities to iteratively transform user intentions into actionable outcomes. While existing methods like CoT~\cite{NEURIPS2022_9d560961} or Reflexion~\cite{NEURIPS2023_1b44b878} improve success rates in regular tasks, they still struggle to effectively handle such complex tasks, with average failure rates over 60\%(see Table \ref{tab:RQ1}). Our study, in alignment with prior research\cite{yang2024buffer}, identifies that the root cause of this problem is that LLMs exhibit constrained generalization capabilities in complex tasks and lack advanced planning mechanisms necessary for tackling the multifaceted demands of complex recommendation problems. This deficiency undermines their ability to adapt effectively to diverse task requirements.} \par

\yu{To address the significant challenges in planning and reasoning posed by the complex mission scenarios described above, we propose a thought-augmented multi-agent IRS named TAIRA. The agents' ability is enhanced by the Thought Pattern Distillation (TPD) method, which distills high-level thought patterns from problem-solving experiences. This process ultilizes multi-scale experiential guidance from both agents and humans, thus enabling agents to handle diverse and complex tasks while achieving strong generalization to novel scenarios. Additionally, we have developed a series of supplementary techniques to further optimize the performance and compatibility of the TPD method. To assess the proposed methods, we conducted a variety of experiments that explore different aspects of the approach, including comparisons with baseline methods, assessments of the impact of various modules, and evaluations of the model's generalization performance. Specifically, the primary contributions of this paper are as follows:}
\yu{\begin{itemize}[leftmargin=*,align=left]
\item We propose and study scenarios involving diverse and complex intents in the interactive recommendation and introduce a user simulation method designed to generate batches of queries for exploring such challenging scenarios.
\item We propose TAIRA, a multi-agent system that decomposes complex intents and provides better recommendations by dynamically planning actionable subtasks.
\item We enhance TAIRA by designing a novel method named Thought Pattern Distillation (TPD), which leverages high-level experiential guidance from both agents and human experts to enhance the agent's reasoning and planning abilities.
\item We conduct comprehensive experiments on multiple datasets; the results from TAIRA demonstrate its significantly enhanced performance compared to existing methods. Notably, TAIRA shows a greater advantage on more challenging tasks while generalizing effectively on novel tasks.
\end{itemize}}

\section{Related Work}
\subsection{Interactive Recommender System}
Interactive Recommender Systems (IRS) enhance user engagement and improve recommendation accuracy by actively involving users through interactions such as ratings, clicks, and natural language input~\cite{HE20169}. While structured and controllable, early attribute-based approaches~\cite{lei2020estimation,lei2020interactive,zhou2020leveraging,zou2020towards} lack the flexibility to handle complex or ambiguous user needs.\par
With advances in machine learning, Conversational Recommender Systems (CRS) have become mainstream within the domain of IRS. CRS engages users through multi-round natural language interactions~\cite{gao2021advances}. They dynamically generate responses based on user input, adopting an end-to-end framework that integrates recommendation and dialogue generation~\cite{tu2022conversational,zhou2022c2,zhou2020improving}, handling multi-turn interactions~\cite{wang2022towards,li2018towards}. However, challenges remain in addressing complex user needs due to limitations in natural language understanding and dialogue context comprehension.\par
The rise of LLMs has fueled significant advancements in recommender systems~\cite{wu2024survey,wang2025generative,guo2024scaling}, including more flexible IRS capable of deeper understanding of such complex or ambiguous input. Unlike rule-based simulations, LLMs capture latent user preferences and behavioral patterns~\cite{liu2023user}, enabling better simulations. Leveraging models like GPT-4, these systems better understand user inputs and generate more personalized dialogues~\cite{10.1145/3626772.3657669,huang2024recommenderaiagentintegrating,shen2025genkienhancingopendomainquestion}.\par
In the context of Interactive Recommender Systems (IRS), intent recognition plays a crucial role in understanding and responding to user needs during dialogue-based interactions. Early methods relied on traditional machine learning techniques~\cite{10.1145/3443279.3443308,10.4108/eetsis.v10i2.2948}, which later evolved into neural network approaches such as LSTM and RNN~\cite{firdaus2021deep,9860839,RIZOU2023200247,chandrakala2024intent}. For example, Sun et al.~\cite{sun2018conversational} used a belief tracker and a deep policy network to guide recommendation actions based on user intent. The advent of LLMs like BERT and GPT shifted the focus to fine-tuning pre-trained models for intent recognition tasks~\cite{vulic2021convfit,robotics13050068,li2024incorporatingexternalknowledgegoal,10.1145/3477495.3532069,app12031610}.  Recently, LLMs such as ChatGPT have demonstrated zero-shot capabilities in intent recognition~\cite{song2023largelanguagemodelsmeet,bodonhelyi2024userintentrecognitionsatisfaction}, further enhancing IRS by reducing reliance on large labeled datasets. Moreover, prompt-driven approaches have emerged, enabling dynamic adaptation to conversational contexts without extensive re-training~\cite{feng2023largelanguagemodelenhanced,10.1007/978-3-031-44693-1_3,mao2023largelanguagemodelsknow}.


\subsection{Agent Planning Enhancement}
\yuold{Several methods can enhance planning in LLM-powered agents, with task decomposition and reflection and refinement being prominent approaches~\cite{huang2024understandingplanningllmagents}. Task decomposition simplifies complex tasks into sub-tasks, as seen in methods like CoT~\cite{NEURIPS2022_9d560961} and Plan\&Solve~\cite{wang2023planandsolvepromptingimprovingzeroshot}. However, these methods can be rigid in dynamic environments, becoming inefficient when unexpected changes occur. Reflection and refinement enhance adaptability by revising plans after failures, as in Reflexion~\cite{NEURIPS2023_1b44b878}, an improvement on ReAct~\cite{yao2022react}, but are less effective in real-time situations. The agent may struggle to correct all errors independently. A more flexible approach is using task experience, like BoT~\cite{yang2024buffer}, which leverages high-level insights from prior problem-solving to enhance real-time thought processes.}

To handle complex user intents, the recommender agent system needs to not only generate efficient task decomposition plans but also ensure that the execution of subtasks can dynamically adapt to uncertainties and changes during task progress~\cite{gu2025rapid}. To achieve this, the recommender agent system needs a reliable planning method. There are two methods of agent planning: Decomposition-First and Interleaved \cite{huang2024understandingplanningllmagents}. The Decomposition-First method, exemplified by Plan-and-Solve~\cite{wang2023planandsolvepromptingimprovingzeroshot}, generates a complete plan at once, creating strong associations between subtasks and the original task. In contrast, the Interleaved method, exemplified by ReAct~\cite{yao2022react}, can dynamically decide the next plan based on the feedback from subtask execution, allowing for the completion of more complex tasks.
However, a single predetermined plan may not be sufficient to address the uncertainties and changes encountered during actual operations. To address this, we adopted a Hierarchical Planning approach that combines unified plan guidance with operational flexibility, ensuring that we can efficiently respond to challenges across different scenarios.

\section{The TAIRA System}

In this section, we formulate the main problem we study. Next, we propose TAIRA and describe each of its components.
Finally, we introduce a thought pattern distillation (TPD) method for enhancing the planning capabilities of the system.


\begin{figure*}[t]
    \centering
    \includegraphics[width = .85\textwidth]{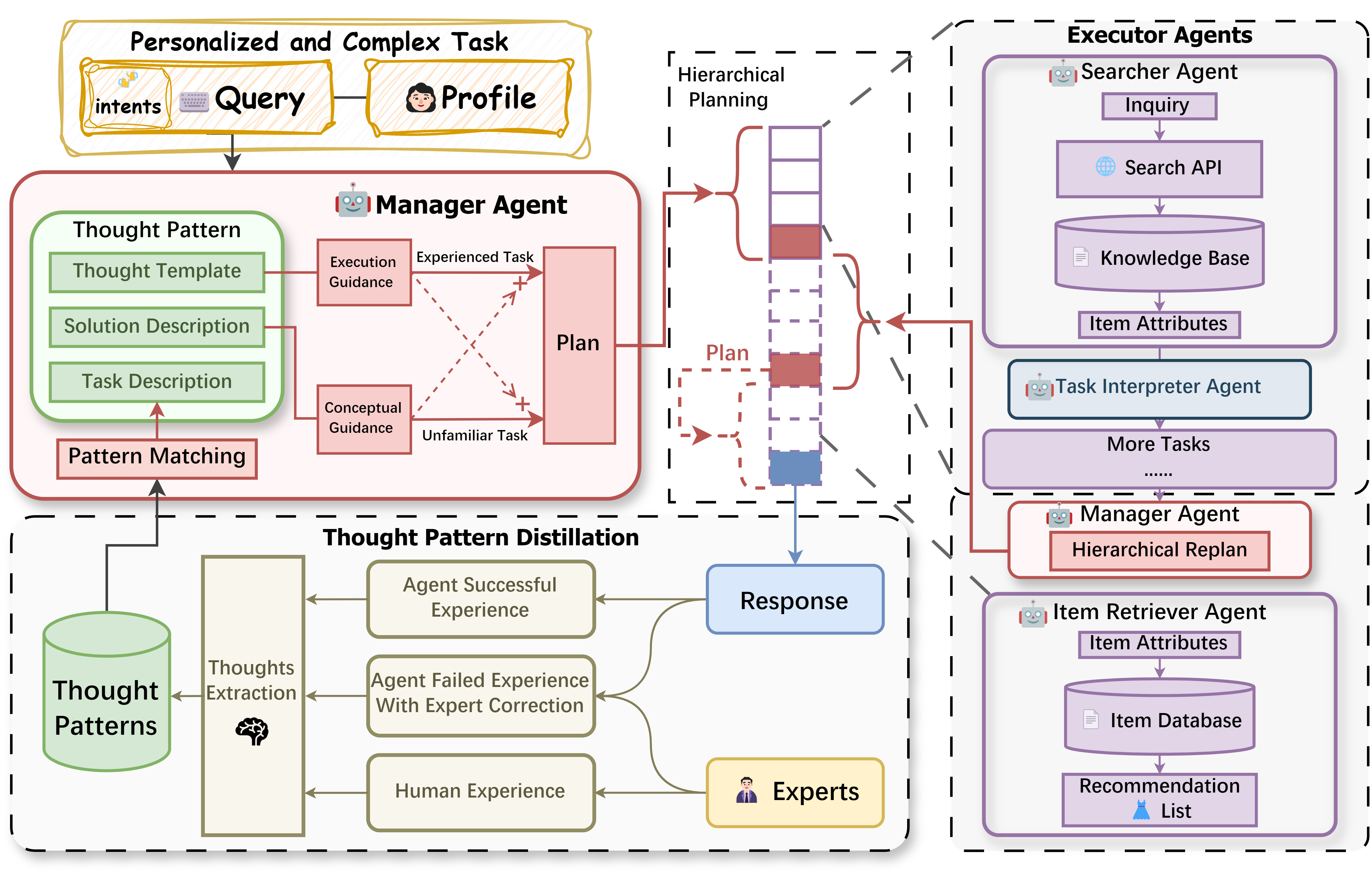}
    \Description{A diagram showing the architecture of TAIRA, consisting of various layers and components.}
    \caption{TAIRA’s overall architecture. The Manager Agent locates a matching Thought Pattern for each user query, then decomposes and plans tasks in phases. Executor Agents complete these sub‐tasks. Meanwhile, Thought Pattern Distillation collects and refines successful, corrected, and expert experiences into thought patterns.}
    \vspace{-1.\baselineskip}
    \label{fig:agentic-workflow}
\end{figure*}

\subsection{Preliminary}

\yu{We study the interactive recommendation task by modeling a user interacting with a recommender system across various levels of intent complexity. The main problem we aim to solve is enhancing agents' reasoning and planning abilities to iteratively transform user intentions into actionable outcomes in recommendation environments. Specifically, we focus on enabling agents to decompose queries with complex intents and strategically plan subtasks to address them by leveraging multi-scale experiential guidance from both agents and humans, while ensuring generalization to novel scenarios.}
Here, we mainly focus on scenarios involving single-turn user input due to the limitations of the available datasets. Indeed, we believe that our formulation with multi-agent architecture can also be generalized with multi-turn user inputs. We leave this as an interesting future work.

\subsection{The Overview Architecture}

\wuold{In interactive recommendations, a user provides a natural-language query with various levels of intent complexity and a recommender agent system recommends items to satisfy the user's information needs.
Figure~\ref{fig:agentic-workflow} shows our proposed thought-augmented interactive recommender agent system (TAIRA) that enhances planning capabilities to address complex user intents by leveraging a series of high-level thought patterns derived from the problem-solving experiences of both agents and humans.
In particular, TAIRA is formulated as an LLM-powered multi-agent system that consists of three main modules:  Manager Agent, Executor Agents, and Thought Pattern \wuold{Distillation}.
The thought-augmented Manager Agent plays a central role by decomposing users' needs and coordinating multiple Executor Agents to effectively perform recommendation tasks.
Specifically, the Manager Agent analyzes the user's query and searches for a suitable Thought Pattern similar to the current one. 
Guided by these high-level thought patterns, the Manager Agent generates a phased plan, which ends either by generating a response or creating the next phase of the plan.
According to the sub-tasks arranged in the plan, the Manager Agent calls Executor Agents to execute the sub-tasks and obtain the corresponding results. 
If further plan generation is scheduled, the next plan is generated; otherwise, a response is generated for the user.
Moreover, we apply a Thought Pattern Distillation (TPD) method to extract reusable thought patterns from the past successful experiences of agents, along with agent failed experiences after human experts' correction, and direct experiences of human experts.}

\subsection{Thought-Augmented Manager Agent}

The Manager Agent is the core module of our proposed TAIRA system, augmented with Thought Patterns to enhance its planning capabilities.
\yu{The so-called \textit{Thought Pattern} is a high-level distillation of the experiences of both agents and humans which provide multi-scale guidance (we will give more details on Thought Pattern Distillation in Section~\ref{subsect:TPD}).}

\paragraph{Thought Pattern Matching.}
\yu{Thought Pattern Matching mirrors how humans recall similar experiences when solving problems. Given a user query, the Manager Agent retrieves the top-K (e.g., top-5) thought patterns by assessing the similarity between the user query and the task descriptions associated with the thought patterns. Subsequently, the Manager identifies the most relevant thought pattern by further comparing the user's query with the task descriptions of each retrieved thought pattern.
}

\paragraph{Hierarchical Planning.}

\wuold{The matched thought pattern is integrated into the Manager Agent's prompt to enhance further task planning.
To address complex user intents, a thought template in the pattern typically consists of multiple phases, with each phase comprising several execution steps.}
To this end, we adopt a \textbf{Hierarchical Planning} approach to update the plan for each phase considering the execution steps in the current phase and the relevant information from the previous phase.

The planning process is essentially an iterative optimization process. Let the initial plan be $P_0$. In each iteration, if the current plan $P_i$ and information set $I_i$ do not fully meet the task requirements, the system will generate a new plan $P_{i+1}$ through the Hierarchical Planning $H$. Specifically, this process can be represented by the following formula:
\begin{equation}
P_{i+1} = H(P_i, I_i), \quad i \geq 0
\end{equation}
This iterative process continually incorporates new information or environmental feedback, progressively refining the plan until a final plan to complete the task is generated.

\subsection{Executor Agents}
To execute the planned subtasks accurately and efficiently, we adopt several Executor Agents to tackle the subtasks accordingly. 

\paragraph{Searcher Agent}

\yu{The Searcher Agent aims to seek relevant knowledge based on the query. When a user asks ``Can you suggest some women’s blouses for a gathering with friends?'', the Search Agent receives an inquiry from the Manager Agent and provides several item attributes, such as ``Stylish'' or ``Casual''.
Equipped with tools like the Google Search API and an Attribute Mapper, it retrieves information from the internet and matches attributes from the domain knowledge base\cite{zhu2024knowagent}. These attributes help efficiently filter items from the candidate pool.
}


\paragraph{Item Retriever Agent}

\yu{The Item Retriever Agent aims to filter and sort the items, and finally provide a ranking list of recommendation items based on the query.
Equipped with a dense retriever, such as BGE-Reranker~\cite{chen2024bge}, it matches the relevant items from the candidate pool according to the similarity between the candidate description and the input attribute keywords.}


\paragraph{Task Interpreter Agent}
\yu{The Task Interpreter Agent translates subtask descriptions into input formats for executor agents. It retrieves task history at the start, combines it with the current subtask description, and generates the input. This enables the multi-agent system to merge static planning with real-time updates.}


\subsection{Thought Pattern Distillation}~\label{subsect:TPD}

\yu{To enhance the agent's reasoning and planning capabilities by leveraging prior experience}, we apply a Thought Pattern Distillation (TPD) method to extract high-level thoughts from the agent's and human experts' experiences, inspired by~\cite{yang2024buffer}. 
Different from~\cite{yang2024buffer}, our proposed Thought Pattern Distillation (TPD) method considers not only the successful experiences of agents but also incorporates corrections from failed experiences and experts' insights to form the thought patterns, \yu{for agents often struggle to complete more complex tasks independently when faced with intricate user intents. In addition, our Thought Pattern also includes guidance at multiple scales to cope with diverse scenarios. }
As shown in Figure \ref{fig:agentic-workflow}, there are three types of sources for successful experiences:

(1) \textbf{Agent Successful Experiences}. 
\wuold{Typically, agents possess certain capabilities to recognize user intents, decompose and plan tasks, and ultimately provide satisfying recommendations. These successful experiences can be automatically distilled into thought templates, enabling the reuse of successful strategies.}

(2) \textbf{Agent Failed Experiences with Expert Correction}. 
\wuold{In this case, an agent may fail to complete a task due to specific steps, and a human expert's correction of its execution route can be considered a successful experience.}

(3) \textbf{Human Expert Experiences}.
\wuold{In this case, users' extensive knowledge and experience in task planning and execution can be effectively distilled into thought templates for agent planning.}

\yuold{Figure \ref{fig:Thought Pattern Extraction} shows the prompt of Thought Pattern Distillation. This prompt instructs an agent how to generate a thought pattern based on the task route, expert's correction (optional), and old pattern.}

\yu{An extracted thought pattern is structured into three components: the task description enables the agent to identify and select the most appropriate thought pattern; the solution description provides conceptual-level guidance for approaching the task; the thought template offers detailed, execution-level instructions, allowing for more fine-grained guidance. As shown in the Manager Agent part of Figure \ref{fig:agentic-workflow}, when a matching thought pattern is found (experienced task), the agent refers to the solution description and follows the steps outlined in the thought template. If no matching thought pattern is available (i.e., a new task is encountered), the agent seeks higher-level guidance mainly from similar solution descriptions to attempt to complete the task. This multi-scale structure of thought patterns provides TAIRA with a foundational level of generalization, enabling it to handle novel tasks without direct experience and maintain a basic success rate without requiring the recording of numerous thought patterns. 
Appendix \ref{appendixB} shows some examples of the distilled thought patterns.}

\begin{figure}[tb]
    \centering
\begin{tcolorbox}[colback=gray!10!white, colframe=gray!70!black, boxsep=0.5mm]
\footnotesize
You are a thought pattern updater and specialize in refining cognitive processes to enhance performance.\par
The thought pattern is a high-level idea extracted from the task execution path, which contains the following three parts: \par
\textbf{1. Task Description}: Used to characterize a task, it is abstract and general, does not include any specific task information, and only describes what type a task belongs to.\par
\textbf{2. Solution Description}: Used to describe the conceptual-level guidance of solving a problem, it is more like the key to solving a class of problems told by a domain expert.\par
\textbf{3. Thought Template}: A template for a path to complete a task, used to indicate the generation of a plan.\par
Based on the given task route and expert opinion, assess how to adjust the old pattern: \{old pattern\}.\par
\textbf{Task Route}: \{task route\}\par
\textbf{Expert Opinion}: \{expert opinion\}\par
The output should align with the structure and tone of the old pattern, maintaining clarity and effectiveness.\par
Provide the revised thought pattern as a concise, actionable statement that improves upon the old approach.
\end{tcolorbox}
\vspace{-1.\baselineskip}
    \caption{The prompt of Thought Pattern Distillation (TPD).}
    \label{fig:Thought Pattern Extraction}\vspace{-1.\baselineskip}
\end{figure}

\begin{table*}[htbp]
	\centering
	\caption{\yu{Examples of different aspects of complexity in the Query construction.}}
 \label{tab:query_cons}\vspace{-1.\baselineskip}
	\resizebox{0.95\textwidth}{!}{
	\begin{tabular}{l|cp{12cm}cc}
    \toprule
	 \textbf{Type of Complexity}& \textbf{Description}& \textbf{Example}& \textbf{Difficulty}& \textbf{Quantity}\\
	\midrule
 	Direct reference& Describe the product directly.& I am looking for a women's pajama set that is thermal and comfortable for the colder months.& easy
& 33.2\%\\
\midrule
 	Occasions& Propose usage occasion.& I want women's sandals that are suitable for beach outings and relaxed gatherings.& medium
& 13.6\%\\
\midrule
 Matching& Additional matching requests.& Additionally, recommend an additional pair of pants and socks to go with these shoes.& medium&11.2\%\\
 \midrule
        Multi Types& Requirements for different types of products.& I'm searching for some classic rock tracks to energize my morning workout. Please recommend me some different styles of music.& medium& 9.7\%\\
\midrule
 Bundle& Request for a bundle of products.& Could you guide me to a set of clothes that are ideal for beach outings and casual gatherings?& hard&10.7\%\\
\midrule
 Ambiguous& With ambiguous requirement.& Could you recommend some women's sneakers that are great for walking and hanging out with friends? I'm not sure about the specific wearing scene.& hard
&11.5\%\\
\midrule
 Multi Occasions& Requirements for different occassions.& Could you suggest a playlist that fits my need for introspective moments during a late-night drive? I'm also looking for something that can help with emotional expression.& hard&10.1\%\\

	\bottomrule
\end{tabular}}\vspace{-1.\baselineskip}
\end{table*}
\section{Construction and Simulation of User Interactions}
In this section, we introduce methodologies for constructing personalized user queries and simulating user behavior to evaluate the recommendations in complex and diverse interaction scenarios.

\subsection{Query Construction}
\yu{To simulate complex and diverse user interaction environments, we developed a series of methods to generate personalized user queries based on recommendation datasets. By analyzing user interaction histories in the dataset, we extract basic information and preferences to construct detailed user profiles. Following the common practice in sequential recommendation modeling~\cite{yin2024dataset}, the last interaction in a user's history is designated as the target item. Subsequently, an LLM generates an initial atomic query from the user's perspective, leveraging information pertaining to this target item~\cite{yin2024dataset, zhang2025td3}. Building on this, we proposed multiple recommendation scenarios with three difficulty levels and introduced different user questioning semantics, inspired by \cite{Saha_Khapra_Sankaranarayanan_2018} and \cite{Wu_2021_CVPR}. The "simple" level involves clear and direct item requests\red{, such as direct reference of products}, while the "medium" level requires reasoning and the conversion of actual needs\red{, such as specifying the occasions of usage}. The "difficult" level typically demands multiple rounds of knowledge acquisition and planning, involves multiple recommendation target, and requires the agent to possess advanced reasoning and planning capabilities. The specific scenarios and difficulty levels are shown in Table \ref{tab:query_cons}, and different semantics are shown in Table \ref{tab:semantics} in the Appendix.
The LLM then classifies each query into the most appropriate scenario and constructs the final query based on the user profile and item information. To simulate a realistic and complex user interaction environment, some scenarios include clear and specific item attributes, while others only present usage requirements, or even ambiguous ones. This approach reflects both explicit and implicit expressions. It is important to note that this method cannot fully encompass all possible real-world scenarios; instead, it aims to generate a rich and representative set of queries that closely resemble the real users' behaviors.
}
\subsection{User Simulation}
\yu{To achieve intelligent and efficient quantitative evaluation, we developed an LLM-driven user behavior simulation inspired by \cite{Wang_2023} and \cite{Simulator}, to evaluate the effectiveness of recommendations for queries constructed. The user's query, profile and the target item are prompted into the simulator. \red{By carefully crafting prompts tailored to specific evaluation scenarios (details available in Figure \ref{fig:user simulator prompt} in appendix), we enable the LLM to emulate realistic user behaviors and preferences in a controlled and consistent manner. The advanced language understanding and generative capabilities of LLM further ensure the reliability of user simulation. The ability of user simulators to model the behavior and needs of different types of users allows flexibility in assessing the quality of recommendations for various scenarios. Note that, there is still a gap between user simulators and real users of interactive recommender systems. Nonetheless, user simulators greatly improve the efficiency and cost of evaluation while ensuring a certain level of reliability.}}

\section{Experimental Setup}

In this section, we evaluate the effectiveness of our proposed TAIRA system by benchmarking it against existing state-of-the-art approaches documented in the literature. The previous section has prepared the data and evaluation methods. 
\wuold{In particular, we aim to address the following four research questions.}

\begin{itemize}[leftmargin=*,align=left]
    \item RQ1: Does our proposed TAIRA approach with thought augmentation outperform the existing state-of-the-art baseline approaches in the interactive recommendation task? 
    \item RQ2: \wuold{How does TAIRA perform on datasets with varying levels of difficulty?}
    \item RQ3: How do the components designed for thought augmentation in TAIRA affect the performance, such as thought pattern matching, hierarchical planning, and agents' and experts' experiences in thought pattern distillation?
    \item \yu{RQ4: How well does TAIRA generalize to novel tasks without direct experience? }
\end{itemize}

\subsection{Datasets \& Setup}
\yu{We applied the designed user query modeling method in Section 4.1 to construct data for experiments on three well-known datasets: \textit{Amazon Clothing \& Shoes} (hereafter referred to as Amazon Clothing), \textit{Amazon Beauty}, and \textit{Amazon Music}~\cite{amazonreview}. 
The Amazon dataset contains extensive information, including product descriptions, categories, related items, user reviews, and interaction history. From this, we selected specific products and their attributes for recommendation, as detailed in Table \ref{tab:datasets} in the appendix. Subsequently, we constructed user queries based on the user and product data at three levels of difficulty. For each difficulty level in each dataset, we selected more than 100 queries. A subset was selected as a development set, with the remainder designated for the testing set. 
The specific quantities are detailed in Table \ref{tab:queries}.}\par
\red{The default LLM used in the experiment is GPT-4o. Additionally, to assess the model's generalization capabilities, we construct a task that TAIRA has not previously encountered by removing the Thought Pattern associated with a specific type of interaction scenario. This task is defined as a \textit{novel task}.}

\subsection{Evaluation Strategies}
\yu{We utilize the user simulator introduced in Section 4.2 to evaluate the effectiveness of our proposed TAIRA and baseline methods across three datasets. \red{The simulator first provides a global assessment, evaluating whether the recommended items align with the user's preferences. It then performs a detailed analysis of each item, assigning additional relevance if the target item is hit, reflecting its suitability. In addition, to better reflect users' personalized preferences, the user simulator incorporates penalties for recommendations that do not align with those preferences.} }\par
To assess performance, we employ top-heavy metrics, such as Hit Ratio (HR) and Normalized Discounted Cumulative Gain (NDCG), along with a task-oriented metric, the Success Rate (SR). 

\begin{itemize}[leftmargin=*,align=left]
    \item \textbf{Success Rate (SR)}~\cite{10.1145/3453154}: 
    We identify two primary failure cases in recommendation tasks. The first occurs when the number of iterations for a subtask exceeds a predefined threshold, suggesting that the subtask is either invalid or redundant. The second arises when there are insufficient items to satisfy user preferences, potentially due to incomplete recommendation coverage or the perception of items as irrelevant.
    \item \textbf{Hit Ratio (HR)}~\cite{10.1145/3269206.3271776}: 
    We consider the hit rate as the frequency of items in the recommendation list that meet the user's needs. The hit rate reflects the extent to which specific recommended items are accepted by users.
    \item \textbf{NDCG}~\cite{10572486}: 
    We adopt Normalized Discounted Cumulative Gain (NDCG) to evaluate the ranking quality of the recommendation results. Items that are relevant and ranked higher are given more weight. NDCG reflects the quality of the recommendation list.
\end{itemize}
\begin{table}[t]
	\centering
	\caption{Queries' statistics.}
 \label{tab:queries}\vspace{-1.\baselineskip}
	\resizebox{0.4\textwidth}{!}{
	\begin{tabular}{l|cc|cc|cc} 
	\toprule
    & \multicolumn{2}{c|}{Amazon Clothing} 
	 & \multicolumn{2}{c|}{Amazon Beauty} &  \multicolumn{2}{c}{Amazon Music} \\
	 & Dev & Test & Dev & Test & Dev & Test \\
	\midrule
 	Easy & 20& 119& 20& 129& 20& 124
\\
 	Medium & 20& 131& 20& 135& 20& 119
\\
        Difficult & 20& 108& 20& 131& 20& 124
\\
	\bottomrule
\end{tabular}}\vspace{-1.\baselineskip}
\end{table}

\begin{table*}[htbp]
\def\ttestB{$\dagger$}
	\centering
	\caption{The effectiveness of the tested approaches. The best results of baselines and the best overall results are underlined and highlighted in bold, respectively. \% Improv. indicates the improvements by our TAIRA approach over the best baseline model. * denotes a significant difference in terms of paired t-test ($p<0.05$), compared to TAIRA.}\vspace{-.5\baselineskip}
	\label{tab:RQ1}
	\resizebox{.85\textwidth}{!}{
	\begin{tabular}{l|ccc|ccc|ccc} 
	\toprule
	 & \multicolumn{3}{c|}{Amazon Clothing}& \multicolumn{3}{c|}{Amazon Beauty}&  \multicolumn{3}{c}{Amazon Music}\\
	Methods & HR@10&NDCG@10& SR & HR@10&NDCG@10& SR & HR@10&NDCG@10& SR \\
	\midrule
        BM25 & 0.0801*& 0.2543*& 0.4929*& 0.2171*& 0.4546*& 0.4886*& 0.1628*& 0.3336*& 0.3910*\\
        BGE-M3 & 0.3103*& 0.4356*& 0.5740*& 0.2694*& 0.4889*& 0.4851*& 0.1187*& 0.2947*& 0.4234*\\
        BGE-Reranker & 0.2970*& 0.5031*& 0.6095*& 0.2980*& 0.4946*& 0.5152*& 0.1763*& 0.3516*& 0.4687*\\
        \midrule
        Zero-shot & 0.3631*& 0.5474*& 0.5809*& 0.2814*& 0.4560*& 0.5625*& 0.1550*& 0.2674*& 0.3916*\\
        One-shot & 0.3705*& 0.5752*& 0.6496*& 0.3491*& 0.5403*& 0.6463*& 0.2287*& 0.4272*& 0.5714*\\
        CoT & 0.3686*& 0.6160*& 0.6857*& 0.3466*& 0.5505*& 0.6333*& 0.2494*& 0.4351*& 0.5448*\\
        Plan \& Solve& 0.4050*& 0.6855 & \underline{0.7966} & 0.3244*& 0.5265*& \underline{0.6573*}& 0.3190*& 0.4857*& 0.6219*\\
        ReAct & 0.4217 & 0.6912 & 0.7231*& 0.3921 & 0.6012 & 0.6034*& 0.2323*& 0.4012*& 0.4713*\\
        Reflexion & \underline{0.4218}& 0.7002 & 0.7750*& \underline{0.4001} & \underline{0.6203} & 0.6444*& \underline{0.3234}& \underline{0.4900*}& 0.5915*\\
        \midrule
        InteRecAgent & 0.4207 & \underline{0.7035} & 0.6883*& 0.3919 & 0.6136 & 0.6485*& 0.2919*& 0.4585*& 0.6155*\\
        MACRec & 0.4205& 0.6846& 0.6717*& 0.3834& 0.5999& 0.6345*& 0.2634*& 0.4876*& 0.\underline{6335*}\\
        MACRS & 0.3871*& 0.6441 & 0.6206*& 0.3156*& 0.5399*& 0.6211*& 0.2348*& 0.4769*& 0.6103*\\
        \midrule
        TAIRA & \textbf{0.4489} & \textbf{0.7314} & \textbf{0.8824} & \textbf{0.4293} & \textbf{0.6579} & \textbf{0.7569} & \textbf{0.3513} & \textbf{0.5546} & \textbf{0.7483}
\\
 	Improv. & 4.54\%& 3.97\%& 9.72\%& 6.80\%& 5.72\%& 13.16\%& 8.40\%& 11.40\%& 15.34\%\\
    \bottomrule
	\end{tabular}}
     \vspace{-1.\baselineskip}
\end{table*}
\subsection{Baselines}
We evaluate the superiority of our proposed TAIRA method by comparing it with three categories of baseline methods. The first category includes different ranking models, including:
\yuold{\begin{itemize}[leftmargin=*,align=left]
    \item \textbf{BM25}~\cite{10.1145/2682862.2682863}: BM25 is a classic probabilistic ranking model that calculates document relevance for ranking.
    \item \textbf{BGE-M3}~\cite{li2023making}: BGE-M3 is a versatile embedding model that enables dense, and sparse retrieval across inputs of varying lengths.
    \item \textbf{BGE-Reranker}~\cite{chen2024bge}: BGE-Reranker is a model that directly calculates the similarity between a query and document, producing a relevance score without embeddings.
\end{itemize}}

The second category includes various agent planning methods, including:
\begin{itemize}[leftmargin=*,align=left]
    \item \textbf{Zero-shot\cite{wei2022finetunedlanguagemodelszeroshot}}: The large model autonomously decides on Executor Agent invocation based solely on the user query and Executor Agent description, without additional prompts.
    \item \textbf{One-shot\cite{brown2020languagemodelsfewshotlearners}}: Each execution agent is provided with an example of how to be invoked.
    \item \textbf{Plan\&Solve}~\cite{wang2023planandsolvepromptingimprovingzeroshot}: A complete plan is generated before execution.
    \item \textbf{COT}~\cite{NEURIPS2022_9d560961}: Before generating the plan, the Chain of Thought is used to break down tasks into thoughts and actions.
    \item \textbf{ReAct}~\cite{yao2022react}: Reasoning and execution are performed alternately through a loop of Thought, Action, and Observation.
    \item \textbf{Reflexion}~\cite{NEURIPS2023_1b44b878}: An enhancement of ReAct, incorporating a self-reflection mechanism.
\end{itemize}

The third category contains different state-of-the-art interactive recommender agent systems powered by LLMs, including:

\begin{itemize}[leftmargin=*,align=left]
    \item \textbf{MACRec}~\cite{10.1145/3626772.3657669}: The MACRec framework introduces multiple agents, where the Manager oversees task allocation and assigns agents to complete the recommendation task.
    \item \textbf{MACRS}~\cite{fang2024multiagentconversationalrecommender}: MACRS uses a multi-agent system where agents execute tasks and propose responses, with the best response selected as the output. It also incorporates user feedback to address failures.
    \item \textbf{InteRecAgent}~\cite{huang2024recommenderaiagentintegrating}: InteRecAgent introduces an interactive recommendation framework based on tool invocation, and incorporates mechanisms including the dynamic demonstration-augmented task planning to enhance the agent's ability to complete recommendation tasks.
\end{itemize}

There are some other recommender system frameworks based on LLM-Agents, such as~\cite{10.1145/3589334.3645537} and~\cite{10572486}. However, \wuold{they are not comparable in this task, since these systems are not interactive recommender systems or do not support the paradigm of interactive recommendation.}

In addition to the above baseline models, we also investigate variants of ablation studies. Such variants can also act as solid baselines:
 \begin{itemize}[leftmargin=*,align=left]
    \item \textbf{TAIRA w/o Thought Pattern Matching (T)}: The TAIRA w/o Thought Pattern Matching variant does not use the guidance of thought patterns. 
    \item \textbf{TAIRA w/o Hierarchical Planning (H)}: The TAIRA w/o Hierarchical Planning variant does not use the Hierarchical Planning mechanism, but directly generates an execution plan based on the instructions of the thought pattern.
    \item \textbf{TAIRA w/o Experts' Experiences (E)}: The TAIRA w/o Experts' Experiences variant only extracts high-level thinking from the agent's execution without using the experiences of human experts and human corrections with failed agents' experiences.
    \item \textbf{TAIRA w/o Agents' Experiences (A)}: The TAIRA w/o Agent' Experience variant only extracts high-level thinking from the experience of human experts without using the agent's execution. 
\end{itemize}

\section{Experimental Results}
In this section, we implement corresponding experiments to answer the four research questions. 
For each experiment, we give the experimental results and detailed analysis to solve the corresponding research question.

\subsection{TAIRA vs. Baselines (RQ1)}

Table \ref{tab:RQ1} shows the performance of our proposed TAIRA method and three categories of baseline methods across three metrics on three datasets. 
We find that our proposed TAIRA method outperforms all other baselines across all metrics. That's because the guidance of high-level thoughts in the TAIRA system focuses on successful experiences, incorporating both self-feedback and external experiences, thus comprehensively enhancing task planning effectiveness and accomplishing tasks. 
We also observe that in the agent planning baselines, the ReAct method outperforms other planning baselines without ReAct on most metrics. Reflexion, as a self-feedback method built on ReAct, further improves performance on all metrics. In the agent-based interactive recommendation frameworks, InteRecAgent demonstrates higher success rates compared to the other baselines.
In fact, Reflexion introduces feedback based on self-execution experiences, while InteRecAgent incorporates task examples generated by external LLMs. These two baselines introduce high-level thoughts, distinct from simple single-step examples, from different perspectives, underscoring the importance of high-level thoughts in solving tasks with complex intentions.

Concerning RQ1, the results clearly show that our proposed TAIRA approach with thought augmentation outperforms the existing baseline approaches in the interactive recommendation task.
\vspace{-1.\baselineskip}

\subsection{Impact of User Intent Complexity (RQ2)}

\begin{figure*}[tb]
\centering
\subfigure[Amazon Clothing]{
\begin{minipage}[t]{0.32\linewidth}
\centering
\includegraphics[width=2.3in]{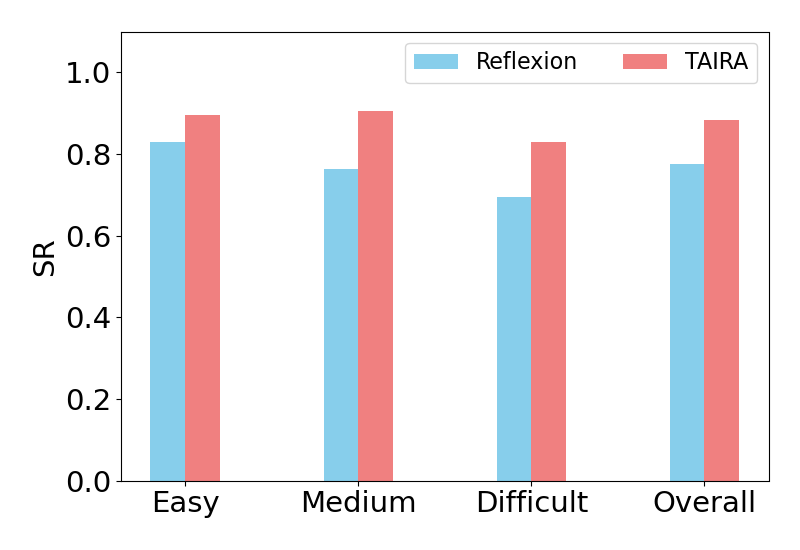}
\end{minipage}%
}%
\subfigure[Amazon Beauty]{
\begin{minipage}[t]{0.32\linewidth}
\centering
\includegraphics[width=2.3in]{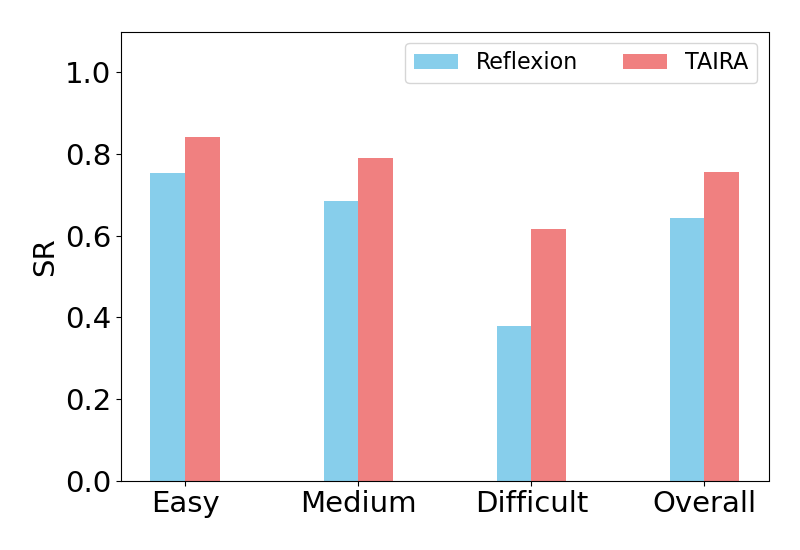}
\end{minipage}%
}
\subfigure[Amazon Music]{
\begin{minipage}[t]{0.32\linewidth}
\centering
\includegraphics[width=2.3in]{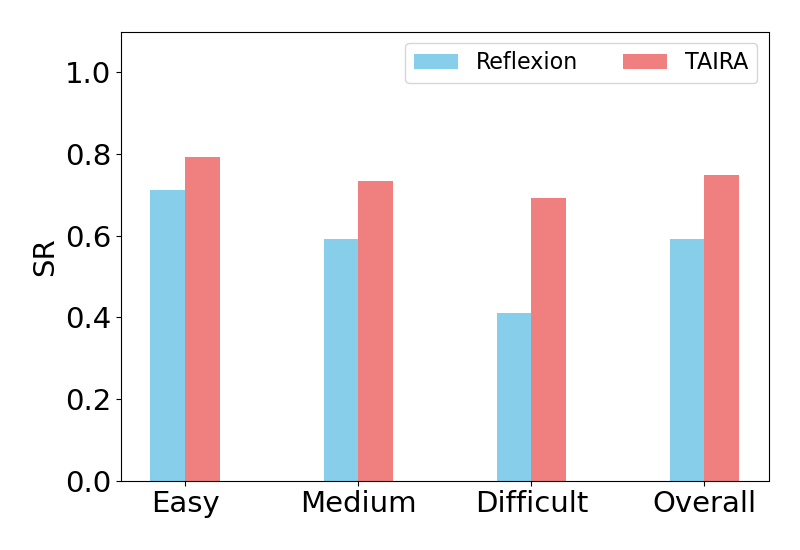}
\end{minipage}%
}
\vspace{-1.\baselineskip}
\centering
\caption{SR of Reflexion and TAIRA across three difficulty levels.}
\label{fig:difficulty} \vspace{-1.\baselineskip}
\end{figure*}

\yuold{To further investigate the improvements of our approach across queries of varying difficulties, we plotted bar charts comparing SR from the TAIRA and Reflexion methods across three difficulty levels in three different datasets, as shown in Figure \ref{fig:difficulty}. 
It can be observed that the improvement of the TAIRA method over the Reflexion method varies depending on the difficulty level. 
For easy queries, the difference between TAIRA and Reflexion is minimal. 
For medium and especially difficult queries, TAIRA demonstrates a clear advantage in all metrics. 
These categories of query contain more complex and ambiguous requirements, which make it difficult for recommender systems to give user-satisfying recommendations without sufficiently reliable guidance from high-level thoughts. 
This indicates that the improvements of TAIRA primarily stem from these more challenging queries, further validating the success of our design for recommendations involving complex intents.}

Thus, in response to RQ2, TAIRA demonstrates superior performance on varying difficulty levels, with particularly significant improvements over the baseline on medium and difficult queries.


\subsection{Impact of Different Components (RQ3)}

\wuold{To address RQ3, Table \ref{tab:ablation} examines the comparative performances of TAIRA without different components, such as Thought Pattern Matching (T), Hierarchical Planning (H), Experts' Experiences (E), and Agents' Experiences (A). }
The results indicate that eliminating any of these mechanisms leads to a decline in performance. Among these mechanisms, removing Thought Pattern Matching (T) has the most significant impact, particularly causing a noticeable decrease in SR. 
This is because Thought Pattern Matching (T) enables TAIRA to avoid failed plans by leveraging successful experiences. 
Removing the Agents' Experiences (A) has a relatively larger impact because, without the execution process as a basis, the Thought Pattern is just an abstract guide without execution. 
Removing Experts' Experience (E) has a smaller impact because the Agent's Experience can guarantee the basic success rate of some tasks. 
The Hierarchical Planning (H) mechanism has the least impact on performance. 
Nonetheless, the performance without these three mechanisms still surpasses that of most baselines. 

\yuold{Therefore, concerning RQ3, the ablation studies confirm that the components designed for thought augmentation in TAIRA all contribute significantly to performance, with Thought Pattern Matching (T) being the most critical.}

\begin{table}[t]
\def\ttestB{$\dagger$}
	\centering
	\caption{Impact of Components. The best results are highlighted in bold. * denotes a significant difference in terms of paired t-test ($p<0.05$), compared to TAIRA.}
	\label{tab:ablation}\vspace{-.5\baselineskip}
	\resizebox{0.45\textwidth}{!}{
	\begin{tabular}{l|cc|cc|cc} 
	\toprule 
	 & \multicolumn{2}{c|}{Amazon Clothing} & \multicolumn{2}{c|}{Amazon Beauty} &  \multicolumn{2}{c}{Amazon Music}\\
  Methods & HR@10& SR & HR@10& SR & HR@10& SR \\
	\midrule
    TAIRA & \textbf{0.4489}& \textbf{0.8824}& \textbf{0.4293}&  \textbf{0.7569}& \textbf{0.3513}& \textbf{0.7483}\\
    \midrule
    1. w/o $T$ & 0.4023*& 0.7935*& 0.3278*& 0.6552*& 0.3176*& 0.6396*\\
    2. w/o $H$ & 0.4080*& 0.8478 & 0.3621*& 0.7143& 0.3486& 0.6809\\
    3. w/o $E$ & 0.4074*& 0.8236& 0.3587*& 0.6812*& 0.3381& 0.6712\\
    4. w/o $A$ & 0.4013*& 0.8081 & 0.3546*& 0.6824*& 0.3484& 0.6592*\\
    \bottomrule
	\end{tabular}} 
 \vspace{-1.\baselineskip}
\end{table}

\subsection{Generalization on Novel Tasks (RQ4)}


\yu{To assess the generalization ability of our proposed TAIRA method on novel tasks, we selected one scenario from each of three difficulty levels and constructed \textit{novel tasks}. We compared the success rates of TAIRA and Reflexion on these tasks, with the original TAIRA as a reference. The results, presented in Table \ref{tab:RQ4}, indicate that TAIRA performs generally worse on \textit{novel tasks} compared to tasks with prior experience. The performance gap is narrow for simpler tasks, but increases with task difficulty. Nevertheless, the success rates for TAIRA remain higher than those for Reflexion, which performs relatively well among the baselines. This improvement is likely due to the conceptual-level guidance provided by the solution descriptions of similar tasks, enhancing TAIRA's ability to generalize. }\par
\yu{Therefore, in response to RQ4, TAIRA demonstrates good generalization on novel tasks for which we have no direct experience. Actually, after the first execution, the novel task will be constructed as a Thought Pattern, further ensuring its usability in applications. }
\vspace{-1.\baselineskip}
\begin{table}[t]
\def\ttestB{$\dagger$}
	\centering
	\caption{\yu{Success rate comparison of TAIRA and Reflexion on \textit{novel tasks}, with the original TAIRA as a reference. The better results are highlighted in bold}}
	\label{tab:RQ4}
	\resizebox{.48\textwidth}{!}{
	\begin{tabular}{l|clc|clc|clc} 
	\toprule
	   & \multicolumn{3}{c|}{Amazon Clothing}& \multicolumn{3}{c|}{Amazon Beauty}&  \multicolumn{3}{c}{Amazon Music}\\
	 Methods & Easy&Medium& Hard& Easy&Medium& Hard& Easy&Medium& Hard\\
	\midrule
         TAIRA& 0.9070&0.8958& 0.8571& 0.9302&0.8302& 0.6111& 0.7931&0.6774& 0.5118\\
	\midrule
         TAIRA on \textit{Novel tasks}& \textbf{0.8973}&\textbf{0.8391}& \textbf{0.7834}&\textbf{0.9172}&\textbf{0.8041}& \textbf{0.5431}& \textbf{0.7713}&\textbf{0.6387}& \textbf{0.5597}\\
        \midrule
         Reflexion & 0.8217&0.7695& 0.7212& 0.7942 &0.7624 & 0.5023 
&0.7123 &0.5734 & 0.4023 
\\
    \bottomrule
	\end{tabular}}
 \vspace{-1.\baselineskip}
\end{table}


\subsection{Case Study}

In Figure \ref{fig:case_TAIRA}, we illustrate the execution path of TAIRA and Reflexion for the given user query. In this case, the user expressed a vague request, seeking recommendations for trousers suitable for gatherings with friends, but did not specify the type or occasion of the gathering, leading to ambiguity in the request. 
Faced with this demand, Reflexion directly searched for information and made recommendations, and after the user's dissatisfaction, did not realize that the failure was due to the biased nature of the recommendation, but instead retrieved more niche items with more detailed criteria, leading to another failure. 
In contrast, our proposed TAIRA method demonstrated stronger understanding and reasoning capabilities than baseline methods. 
Through Thought Pattern matching, the Manager realized that the user's fuzzy requirements need to be processed in stages. 
First, through knowledge acquisition, the types of needs are subdivided. 
On this basis, the steps that need to be performed for each type of recommendation are re-planned.
In the end, the recommendations that cover the user's potential needs are successfully provided, making the user feel satisfied.

\begin{figure}[t]
    \centering
    \begin{tcolorbox}[colback=gray!10!white, colframe=gray!70!black]
        \scriptsize
        \definecolor{darkblue}{RGB}{0, 0, 150}
        \definecolor{darkred}{RGB}{150, 0, 0}
        \definecolor{darkgreen}{RGB}{0, 150, 0}
        \textcolor{darkblue}{\textbf{\small{User query:}}} {\scriptsize{Can you suggest some blouses for a gathering with friends? I'm not sure about the specific wearing scene.}}\par
        
        \setlength{\columnsep}{12pt}           
        \setlength{\columnseprule}{0.4pt}      
        \def\columnseprulecolor{\color{black}} 

        \begin{multicols}{2}
        \textbf{\small{Execution Route of Reflexion}} \par
        \textbf{Thought1:} blouses for a gathering with friends is not a clear attribute of the item and I need to search for information about it.\par
        \textbf{Action1[Searcher]:} Search for attributes of blouses for gathering with friends.\par
        \textbf{Observation1:} Attributes of casual clothing.\par
        \textbf{Thought2:} I have got enough information to retrieve the item.\par
        \textbf{Action2[Item Retriever]:} Retrieve blouses with attributes for a casual gathering as well as the user's preferences.\par
        \textbf{Observation2:} Some women's casual blouses.\par
        \textbf{Thought2:} I have got enough information to make a recommendation.\par
        \textcolor{darkblue}{\textbf{Recommend}} \par
        List 1: blouses with attributes for casual gatherings.\par
        \textcolor{darkred}{\textbf{Fail}}\par
        
        \textbf{Reflection:} The item attributes used for retrieval are not accurate enough.\par

        \textbf{Thought3:} I need a more detailed search.\par
        \textbf{Action3[Searcher]:} Search for specific attributes of blouses for gathering with friends.\par
        \textbf{Observation3:} More detailed attributes.\par
        \textbf{Thought4:} I have to retrieve the item again.\par
        \textbf{Action4[Item Retriever]:} Retrieve blouses with detailed attributes for a casual gathering.\par
        \textbf{Observation4:} Some women's niche blouses.\par
        \textbf{Thought4:} I have got enough information to make a recommendation.\par
        \textcolor{darkblue}{\textbf{Recommend}} \par
        List 1: Women's highly niche blouses with attributes for casual gatherings.\par
        \textcolor{darkred}{\textbf{Fail}}\par
        \columnbreak  
        \textbf{\small{Execution Route of TAIRA}} \par
        \textcolor{darkred}{\textbf{TPD[Manager]:}} Search for a thought pattern, which instructs separate recommendations for different demand scenarios and separate handling of user preferences.\par
        \textcolor{darkred}{\textbf{Plan[Manager]:}} Generate an initial plan for obtaining information, following the guidance of the thought pattern.\par
        \textbf{Task1[Searcher]:} Search for different types of gatherings with friends and the suitable blouse styles for each.\par
        \textbf{Output1}: Types of casual gatherings and semi-formal gatherings + Available attributes for casual gatherings.\par

        \textcolor{darkred}{\textbf{Replan[Manager]:}} Generate a recommendation plan for blouses suitable for casual gatherings and semi-formal gatherings.\par
        
        \textbf{Task2[Searcher]:} Search for blouse styles suitable for semi-formal gatherings with friends.\par
        \textbf{Output2}: Attributes for semi-formal gatherings.\par
        \textbf{Task3[Item Retriever]:} Retrieve items for blouses with attributes for casual gatherings. Then Reorder based on user preference.\par
        \textbf{Output3}: Some women's casual blouses.\par

        \textbf{Task4[Item Retriever]:} Retrieve blouses with attributes for a semi-formal gathering. Then Reorder based on user preference.\par
        \textbf{Output4}: Some blouses for a semi-formal gathering.\par
        \textcolor{darkblue}{\textbf{Recommend}} \par
        List 1: blouses with attributes for casual gatherings.\par
        List 2: blouses with attributes for a semi-formal gathering.\par
        
        \textcolor{darkgreen}{\textbf{Success}}
        
        \end{multicols}

        \setlength{\columnsep}{10pt}  
        \setlength{\columnseprule}{0pt}  

    \end{tcolorbox}
    \vspace{-1.\baselineskip}
    \caption{Executing Case of TAIRA.}
    \label{fig:case_TAIRA}\vspace{-1.\baselineskip}
\end{figure}

\section{Conclusions}
In this paper, we introduced TAIRA, a Thought-Augmented Interactive Recommender Agent system designed to tackle complex and diverse user intents. 
Using the Thought Pattern Distillation method, TAIRA enhances reasoning and planning with multi-scale experiential guidance from past interactions of agents and human experiences, improving both precision and flexibility. 
Experimental results on benchmark datasets show that TAIRA outperforms baseline methods in recommendation accuracy and quality. 
Analysis of its core components and generalization highlights its adaptability to evolving user needs. 
TAIRA marks a significant advancement in interactive recommendation systems with augmented reasoning capabilities. 
Future work will focus on incorporating multi-turn dialogues, enhancing multi-agent collaboration, and extending TAIRA's applications beyond recommendation systems to ensure its robustness and adaptability across diverse real-world scenarios.

\balance

\bibliographystyle{ACM-Reference-Format}

\bibliography{reference}

\appendix
\onecolumn

\section{Prompts}

\subsection{Manager Agent for Planning}
Figure \ref{fig:manager planning} shows the prompt of the manager agent for planning. \yuold{This prompt instructs the Manager Agent how to generate a plan based on the thought template.}
\renewcommand{\thefigure}{A1.1}
\begin{figure}[H]
    \centering
\begin{tcolorbox}[colback=gray!10!white, colframe=gray!70!black ]
You are a manager agent of a conversational recommendation system. You are good at analyzing user inquiry intent and planning tasks. In addition, you are good at transferring the high-level thinking processes of previous successful experiences to current problems.\par
Here are the available agents and their functionalities: \{AGENTS INSTRUCTION\}\par
The user's input is:  “\{ user input\} “.  \par
Based on the user's input, create a task plan in JSON format with sub-tasks. \par
The output should be in JSON format as follows: \par
\{  \par
   “user\_input “:  “\{user input\} “ , \par
   “main\_task “:  “... “ , \par
   “sub\_tasks “: \{  \par
     “task\_1 “: \{ “content “:  “... “ ,  “agent “:  “... “\},  \par
     “task\_2 “: \{ “content “:  “... “ ,  “agent “:  “... “\},  \par
       ...... \par
  \}  \par
\} \par
'Content' is what the agent should do. And 'agent' specifies the agent to execute each sub-task. Remember: PlannerAgent and InteractorAgent **must** be the last sub-task in the plan. No sub-tasks are allowed after a task is assigned to either PlannerAgent or InteractorAgent. You can only use PlannerAgent or InteractorAgent once, and it must be in the final sub-task. There should be no sub-tasks after that. If you think the current task can be completed with a single plan, choose an InteractorAgent, otherwise, choose a PlannerAgent to update the plan after getting enough information. \par
You need to follow the following thinking template to complete the task.\par
This template is a high-level thinking process summarized from the successful experience of similar tasks: \{template\}\par
Among them, solution description is the thinking mode at the level of ideas, while thought template is the thinking mode at the level of execution. \par
You need to judge whether this template is suitable for solving this problem. \par
If it is suitable, you should follow it. If it is not suitable, you can only get inspiration from \par
solution description and imitate the mode of thought template to solve the problem.\par
\end{tcolorbox}
    \caption{The Prompt of the Manager Agent for Planning.}
    \label{fig:manager planning}
\end{figure}

\renewcommand{\thefigure}{A1.2}
\begin{figure}[H]
    \centering
\begin{tcolorbox}[colback=gray!10!white, colframe=gray!70!black ]
The following is the history of tasks executed so far: \{history\}\par
You need to continue to follow the following thinking template to complete the task.\par
This template is a high-level thinking process summarized from the successful experience of similar tasks: \{template\}\par
Among them, solution description is the thinking mode at the level of ideas, while thought template is the thinking mode at the level of execution. \par
You need to judge whether this template is suitable for solving this problem. \par
If it is suitable, you should follow it. If it is not suitable, you can only get inspiration from \par
solution description and imitate the mode of thought template to solve the problem.\par
\end{tcolorbox}
    \caption{The Additional Prompt of the Manager Agent for Hierarchy Planning.}
    \label{fig:manager hierarchy planning}
\end{figure}

\subsection{Executor Agents}
Figure \ref{fig:executor agents} shows the prompt of executor agents' instruction. \yuold{This prompt describes the input, output, and functions of each Executor Agent, and instructs the Manager Agent and Task Interpreter how to call and interpret tasks. The PlannerAgent here is part of the ManagerAgent function. To make it easier for LLM to understand, it is extracted separately. If PlannerAgent is selected at the end, the replan function of ManagerAgent will be called.}
\renewcommand{\thefigure}{A2}
\begin{figure}[H]
    \centering
\begin{tcolorbox}[colback=gray!10!white, colframe=gray!70!black ]
Here are the available agents and their functionalities:\par
- \textbf{ItemRetrievalAgent}: Input a recommendation request containing product attributes (for example:'Please recommend me a sugar-free energy drink.' Such requirement needs to be converted into specific item attributes through SearcherAgent), and recommend a list containing 10 specific items based on keyword similarity. A search by SearcherAgent can only retrieve one target for one requirement. When multiple targets need to be recommended, ItemRetrievalAgent needs to be called multiple times.  \par
- \textbf{SearcherAgent}: Input a short query and search for product attributes that meet the needs from a knowledge base of product attributes and usage, based on keyword similarity. The query target can only be the attributes that meet the target requirements, such as 'what kind of shorts are suitable for mountain climbing'. Note that it is the attributes and not other things. You cannot find other information through this agent. The attributes returned by SearcherAgent are guaranteed to be retrieved in ItemRetrievalAgent. \par
Even if you think they are not appropriate, they are the closest answers to the search input. \par
- \textbf{InteractorAgent}: Generate a final response with one or more recommend result lists (the input does not need to include the recommended items)\par
- \textbf{PlannerAgent}: Input the re-plan goal(for example, The task history provides a list of available product types. Select the two most suitable ones and then enter: 'Generate a recommendation plan for type A and type B'. The types of products must be specific product names, with one recommendation for each type. The number of product types should not exceed 2. And it should not include product types that have been recommended before.) and Regenerate subsequent tasks in the same way as the initial plan based on the information obtained from the executed subtasks. It marks the end of a phased plan. This Agent can **ONLY be placed at the end of a phased plan!!!**
\end{tcolorbox}
    \caption{The Prompt of Executor Agents' Instruction (AGENTS INSTRUCTION).}
    \label{fig:executor agents}
\end{figure}

\subsection{Item Retriever Agent}
Figure \ref{fig:ItemRetriever prompt} shows the prompt of the item retriever agent. \yuold{This prompt instructs the Item Retriever Agent how to obtain user requirements from the input and separate the product and preference information.}
\renewcommand{\thefigure}{A3}
\begin{figure}[H]
    \centering
\begin{tcolorbox}[colback=gray!10!white, colframe=gray!70!black ]
You're a recommendation assistant and you're good at recognizing user preferences.\par
The user's query is:\{user input\}. From this, please extract the user's requirements and preferences for clothing." Please fill in this format and only output the filled content:[clothing type]; [preference]. clothing type is the basic attribute and gender distinction. Other attributes are in preference. The total length must not exceed 15 words. Separate multiple attributes with ' '. Include as many of the key points of user requirements as possible, and the basic attributes of the product are prioritized, followed by the detailed attributes. You only need to reflect the preferences in the user input without making any inferences.
\end{tcolorbox}
    \caption{The Prompt of the Item Retriever Agent.}
    \label{fig:ItemRetriever prompt}
\end{figure}

\subsection{Searcher Agent}
Figure \ref{fig:Searcher prompt} shows the prompt of a searcher agent. \yuold{This prompt instructs the Searcher Agent how to obtain useful information from the search results and summarize it in a standard format.}
\renewcommand{\thefigure}{A4}
\begin{figure}[H]
    \centering
\begin{tcolorbox}[colback=gray!10!white, colframe=gray!70!black ]
You are a searcher agent and you excel at acquiring previously unknown knowledge through search results.\par
Based on the following search results, provide an insight into the target query: \{query\}. 
Search Results: \{context\} The output should only contain specific descriptions. The output is a keyword combination of no more than 20 words, not a descriptive sentence. You should give a specific answer to the question.
\end{tcolorbox}
    \caption{The Prompt of the Searcher Agent.}
    \label{fig:Searcher prompt}
\end{figure}

\subsection{Task Interpreter Agent}
Figure \ref{fig:Task Interpreter prompt} shows the prompt of the Task Interpreter. \yuold{This prompt instructs the Task Interpreter how to generate input into a format acceptable to the next Agent based on the task execution history and the previous output information.}
\renewcommand{\thefigure}{A5}
\begin{figure}[H]
    \centering
\begin{tcolorbox}[colback=gray!10!white, colframe=gray!70!black ]
You are a task planning agent of a conversational recommendation system.\par
You are good at analyzing user inquiry intent and planning tasks.\par
{AGENTS INSTRUCTION}\par
Here is the previous task history:{history str}\par
The current task is “{content}”\par
The next agent to complete this task is: “{next agent name}”. \par
The previous task output is: “{output}”. \par
Based on this information, generate the query for the next agent to make sure it can complete the task and generate the right output.

\end{tcolorbox}
    \caption{The Prompt of the Task Interpreter Agent.}
    \label{fig:Task Interpreter prompt}
\end{figure}

\subsection{Interactor Agent}
Figure \ref{fig:Interactor prompt} shows the prompt of the Task Interactor. This prompt instructs the Interactor how to summarize the results of the task execution and give a final response.
\renewcommand{\thefigure}{A6}
\begin{figure}[H]
    \centering
\begin{tcolorbox}[colback=gray!10!white, colframe=gray!70!black ]
You are a response agent of a conversational recommendation system.
You are good at analyzing provided information and generate recommendation response.\par
Here is the previous task history:\{history str\}\par
Based on the task history, and the instruction from manager:\{instruction\}\par
If you've got enough recommend list, generate a response with one or ore lists, each list containing 10 recommended items (id and title). 
You need to correctly understand the intent in the **complete** task history and include a list of **all** the recommendations needed in the final response. Especially when there are multiple plans for the task execution.\par
Output the lists using the following JSON format:\{json format\}
In the 'recommendation', you should use no more than 5 words to describe the basic type of product you are recommending, especially the product category. Then the 'items' is a list of recommendations for this target. \par
In item information, you must keep as many keywords as possible in the input words when searching for these items. You cannot remove these keywords because they will be used to evaluate the quality of recommendations.\par
You must output 10 items for each list.\par
\end{tcolorbox}
    \caption{The Prompt of the Interactor Agent.}
    \label{fig:Interactor prompt}
\end{figure}

\subsection{User Simulator}
Figure \ref{fig:user simulator prompt} shows the major instruction prompt of the user simulator. \yuold{This prompt instructs the User Simulator how to act as a real user based on the query to understand this complex requirement, and how to evaluate the quality of the recommendations given by the system and give a score.}
\renewcommand{\thefigure}{A7}
\begin{figure}[H]
    \centering
\begin{tcolorbox}[colback=gray!10!white, colframe=gray!70!black ]
You are a shopper who is asking the interactive recommender system for a certain need. \par
You need a product that can truly meet your needs.\par
Your query that contains complete requirements is: \{query\}.\par
There are such sample products that can meet part of your requirement: \{sample product\}.\par
In particular, your requirements in this scenario have the following characteristics: \{scenario description\}\par
Imagine you are in this real-life situation and carefully understand your needs.\par
You will be given one or more 10-item recommendation lists. The items in each list point to the same target.\par
Each listing has a description of the recommended target.\par
First, you need to determine whether these targets together can fully meet your requirements. If not, then the recommendation will be considered a failure.\par
Then, you need to judge whether each of these targets meets your requirements.\par
If the requirements are not met, then this list is considered a failure and all items in it will receive 0 points.\par
Next, you need to determine whether each product meets your needs.\par
The recommendation list for the recommendation target given by the recommendation system is: \{recommendation list\}.\par
Output a list of 10 ratings to express your judgment. \par
The order in the rating list should correspond to the order of the items in the recommendation list. \par
If it meets the requirements, it will correspond to 1 point, if it does not meet the requirements, it will correspond to 0 points. In particular, if it is exactly the same as the sample product, it will be given 2 points. You also need to decide whether each product meets your preferences in some way. Then, among the products that are scored 1, change the score of those that do not meet your preferences to 0.5. 
You should first output your reason, and then output the fail tag and final score lists in a JSON format:'{json format}', the score is a pure number.\par
\end{tcolorbox}
    \caption{The Major Instruction Prompt of the User Simulator.}
    \label{fig:user simulator prompt}
\end{figure}

\section{Thought Pattern Examples}
\label{appendixB}
Figure \ref{fig:thought pattern1}, \ref{fig:thought pattern2}, \ref{fig:thought pattern3}, \ref{fig:thought pattern4} shows examples of part of the thought patterns. 
\renewcommand{\thefigure}{B1}
\begin{figure}[H]
    \centering
    \begin{tcolorbox}[colback=gray!10!white, colframe=gray!70!black]
    \textbf{"template 1:"} \par
        "task description": "\par
        In this conversational recommendation query, the user puts forward one usage requirement and specifies one \par
        clothing type for recommendation.\par
        ",\par
        "solution description": "\par
        To complete the recommendation, you need to generate an execution plan that can successfully generate the recommendation. \par
        In order to complete the recommendation, you need to search for items that meet the target attributes \par
        in the item database through keywords. However, for some complex user needs, the item attributes are unclear. \par
        At this time, you need to call the searcher to search for relevant knowledge and obtain the item attributes \par
        that can meet the user's needs.\par
        ",\par
        "thought template": "\par
        Step 1: Determine the user's target products and needs.\par
        Step 2: Determine whether there are clear product attributes for item retrieval. If not, you need to obtain relevant knowledge through searcherAgent.\par
        Step 3: Use the obtained item attributes to retrieve items in ItemRetriever.\par
        Step 4: Recommend the retrieved items to the user.\par
        "\par
    \end{tcolorbox}
    \caption{Examples of Thought Pattern 1.}
    \label{fig:thought pattern1}
\end{figure}

\renewcommand{\thefigure}{B2}
\begin{figure}[H]
    \centering
    \begin{tcolorbox}[colback=gray!10!white, colframe=gray!70!black]
    \textbf{"template 2":} \par
        "task description": "\par
        In this conversational recommendation query, the user raises a specific demand without specifying \par
        a specific type of apparel, but instead asks for a set of clothes that can meet the demand to be recommended.\par
        ",\par
        "solution description": "\par
        In this task, the user asks for a set of clothing recommendation, but does not specify a specific category. \par
        In order to recommend clothing that satisfies the user, we should first collect information to determine which categories of clothing to recommend, \par
        and then plan the recommendations for each category separately and give their own recommendation lists.\par
        ",\par
        "thought template": "\par
        Phase 1: Collect information to determine which categories of clothing to recommend\par
        Step 1: Determine the user's needs.\par
        Step 2: Based on the need, obtain relevant knowledge through searcherAgent.\par
        Step 3: Based on the knowledge, determine about three clothing types to recommend and update plan.\par
        Phase 2: Provide recommendations for the clothing categories chosen. \par
        Step 1: For clothing type A, determine whether there are clear product attributes for item retrieval. If not, you need to obtain relevant knowledge through searcherAgent.\par
        Step 2: Use the obtained item attributes to retrieve items in ItemRetriever.\par
        Step 3: For the remaining clothing types, follow similar steps.\par
        Step 4: Recommend each final list to the user.
        "\par
    \end{tcolorbox}
    \caption{Examples of Thought Pattern 2.}
    \label{fig:thought pattern2}
\end{figure}

\renewcommand{\thefigure}{B3}
\begin{figure}[H]
    \centering
    \begin{tcolorbox}[colback=gray!10!white, colframe=gray!70!black]
    \textbf{"template 3":} \par
        "task description": "\par
        In this conversational recommendation query, the user puts forward a specific requirement and \par
        asks for a recommendation of a clothing of a specified type, and then requires one or multiple clothing types to be matched with this cloth.\par
        ",\par
        "solution description": "\par
        In this task, the user specify a specific clothing type, and ask for other types to be matched with this product.\par
        In order to recommend clothing that satisfies the user, the recommendation for the specified item should be obtained first, \par
        and then the attributes of the matching clothing should be determined based on the demand and the attributes of the item.\par
        Based on these attributes, recommend these matching clothes.\par
        ",\par
        "thought template": "\par
        Step 1: Determine the user's target product and needs.\par
        Step 2: Determine whether there are clear product attributes for this item retrieval. If not, you need to obtain relevant knowledge through searcherAgent.\par
        Step 3: Use the obtained item attributes to retrieve items in ItemRetriever.\par
        Step 4: The attributes of clothing for matching can be obtained through searcherAgent based on the recommended clothing attributes or the requirements in the query.\par
        Step 5: Use the obtained item attributes to retrieve clothing for matching in ItemRetriever.\par
        Step 6: Recommend each final list to the user.\par
        "\par
    \end{tcolorbox}
    \caption{Examples of Thought Pattern 3.}
    \label{fig:thought pattern3}
\end{figure}

\renewcommand{\thefigure}{B4}
\begin{figure}[H]
    \centering
    \begin{tcolorbox}[colback=gray!10!white, colframe=gray!70!black]
    \textbf{"template 4":} \par
        "task description": "\par
        In this conversational recommendation query, the user puts forward a requirement, \par
        but is not clear about the specific using scene or demand. And asks for a recommendation for a specific type of product.\par
        ",\par
        "solution description": "\par
        In this task, the user puts forward a demand for clothing or beauty product recommendation, but expresses uncertainty about the demand, \par
        which means that more considerations need to be given to meet the needs of different possible specific scenarios.\par
        In order to recommend clothing or beauty product that satisfies users, we should first collect information and determine which \par
        categories the product attributes that meet user needs can be divided into (corresponding to different specific scenarios),\par
        and then plan recommendations for each specific situation and give their own recommendation lists.\par
        ",\par
        "thought template": "\par
        Phase 1: Gather information to determine what categories the target needs can be classified into.\par
        Step 1: Determine the user's needs.\par
        Step 2: Based on the need, obtain relevant knowledge through searcherAgent.\par
        Step 3: Based on the knowledge, divide the user's need into about three specific scenarios. Only one recommendation isn't enough. \par
        Phase 2: Provide recommendations for each specific scenario. \par
        Step 1: For specific scenario A, determine whether there are clear product attributes for item retrieval. If not, you need to obtain relevant knowledge through searcherAgent.\par
        Step 2: Use the obtained item attributes to retrieve items in ItemRetriever.\par
        Step 3: For the remaining product types, follow similar steps.\par
        Step 4: Recommend each final list to the user.
        "
    \end{tcolorbox}
    \caption{Examples of Thought Pattern 4.}
    \label{fig:thought pattern4}
\end{figure}

\section{Experts' Instruction Examples}
\label{appendixC}
Figure \ref{fig:expert_experience}, \ref{fig:expert_correction} show examples of part of the instructions from experts. 
\renewcommand{\thefigure}{C1}
\begin{figure}[H]
    \centering
    \begin{tcolorbox}[colback=gray!10!white, colframe=gray!70!black]
    In this task, the user will specify a particular type of clothing and ask for recommendations of other clothing items that can be paired with it. \par
To meet the user's needs, you first need to identify the recommended item based on their request. \par
Then, consider the attributes of that item and the user's requirements to figure out what kind of clothing would pair well with it. Once you've got that figured out, suggest the matching clothes accordingly.
    \end{tcolorbox}
    \caption{Examples of Expert Experience.}
    \label{fig:expert_experience}
\end{figure}

\renewcommand{\thefigure}{C2}
\begin{figure}[H]
    \centering
    \begin{tcolorbox}[colback=gray!10!white, colframe=gray!70!black]
The reason why recommender failed was that it ignored the user's second sentence: I'm not sure about the specific wearing scene. \par
Since the user is not clear about the specific wearing scene, in order to satisfy the user in one recommendation, 
it is necessary to give recommendations in various possible directions. \par
First, information should be collected to determine 
which categories (corresponding to different specific scenarios) the product attributes that meet the user's needs can be divided into.\par
Then, recommendation planning should be carried out for each specific situation, and a corresponding recommendation list should be given.
    \end{tcolorbox}
    \caption{Examples of Expert Correction.}
    \label{fig:expert_correction}
\end{figure}

\section{Dataset and Query Statistics}
\label{appendixD}
\begin{table}[h]
	\centering
	\caption{\yu{Datasets' statistics.}}
 \label{tab:datasets}\vspace{-1.\baselineskip}
	\resizebox{\textwidth}{!}{
	\begin{tabular}{l|cc|cc|cc} 
	\toprule
    & \multicolumn{2}{c|}{Amazon Clothing} 
	 & \multicolumn{2}{c|}{Amazon Beauty} &  \multicolumn{2}{c}{Amazon Music} \\
	 & quantity& examples& quantity& examples& quantity& examples
\\
	\midrule
 	Item Extracted
& 100K& Sebago Men's Spinnaker Boat Shoe& 100K& Essie Apricot Cuticle Oil Nail Treatment& 300K& All Night Wrong\\
 	attributes
& 943& Shoes | Loafers \& Slip-Ons | Spring Promo | Men | Sebago& 346& Hands \& Nails | Cuticle Care | Skin Care | Cuticle Creams \& Oils& 593& Progressive Metal | Progressive | Rock\\
        meta-info
& 2& 108& 2& 131& 2& 124
\\
	\bottomrule
\end{tabular}}\vspace{-1.\baselineskip}
\end{table}
\begin{table*}[h]
	\centering
	\caption{\yu{Examples of Different Semantics in the Query Construction.}}
 \label{tab:semantics}\vspace{-1.\baselineskip}
	\resizebox{\textwidth}{!}{
	\begin{tabular}{l|p{12cm}c}
    \toprule
	 Type of Semantic& Example& Quantity\\
	\midrule
 	Can you recommend& Could you recommend a bodysuit that fits for family gatherings?& 18.5\%\\
\midrule
 	I am looking for& I am looking for a women's pajama set that is thermal and comfortable for the colder months.& 16.6\%\\
\midrule
 Do you have any suggestions & Do you have any suggestions for a men's T-shirt that's good for music festivals?&15.5\%\\
\midrule
 Can you help me choose& Can you help me choose a music collection that captures the lively atmosphere of a French Quarter street performance?&14.2\%\\
\midrule
 What's the best& What's the best pumps I can use for a formal wedding?&10.5\%\\
\midrule
 Show me& Show me a men's sandals that work well for family beach outings.&9.2\%\\
 \midrule
 I need advice on choosing& I need advice on choosing a cover-up for sun protection during tropical vacations.&8.9\%\\
\midrule
        Where can I find& Where can I find a Disney children's music album with educational and entertaining songs for kids?& 6.6\%\\
	\bottomrule
\end{tabular}}\vspace{-1.\baselineskip}
\end{table*}

\twocolumn

\section{Experimental Supplements}
\subsection{Efficiency Analysis}
The efficiency of systems based on LLM, especially multi-agent systems, is a widely discussed topic. This section will discuss the execution efficiency of TAIRA and provide a theoretical analysis.

The execution overhead of TAIRA mainly manifests in two areas: the tool invocation layer of the Executor Agents and the planning layer of the Manager Agent. Compared to baseline methods, TAIRA does not significantly increase the number of LLM invocations. The analysis is presented from the following two perspectives:

Equivalence of tool invocation layer overhead: The Executor Agents in the TAIRA framework (such as Searcher and Item Retriever) are almost equivalent in terms of efficiency for single tool executions compared to the tool invocation modules in the baseline methods, without introducing significant additional overhead.

Limited planning layer overhead: TAIRA introduces additional elements like thought pattern matching and enhanced thinking templates in the planning prompts, which slightly increase the number of tokens. However, the output format remains consistent with the baseline methods. More importantly, through the enhancement of its thought patterns, TAIRA effectively reduces the number of ineffective tool invocations compared to some baseline methods, thus somewhat balancing the increase in input tokens.

\end{document}